\documentclass[10pt,twocolumn,letterpaper]{article}

\usepackage[pagenumbers]{cvpr}
\usepackage[dvipsnames]{xcolor}
\usepackage[permil]{overpic}

\definecolor{cvprblue}{rgb}{0.21,0.49,0.74}
\definecolor{blendifyblue}{rgb}{0.14,0.35,0.53}
\definecolor{blendifygray}{rgb}{0.35,0.35,0.35}

\newcommand{\blendify}[0]{{\fontfamily{lmss}\selectfont
{\color{blendifyblue}blend}{\color{blendifygray}ify}}}

\def\cput(#1,#2)#3{%
  \put(#1,#2){\fontfamily{lmss}\selectfont{#3}}%
}

\newcommand\blfootnote[1]{%
  \begingroup
  \renewcommand\thefootnote{}\footnote{#1}%
  \addtocounter{footnote}{-1}%
  \endgroup
}

\newcommand{\PAR}[1]{\vskip4pt \noindent{\bf #1~}}

\def\code#1{{\fontfamily{courier}\selectfont
{#1}}}

\makeatletter
\newcommand*{\rom}[1]{\expandafter\@slowromancap\romannumeral #1@}
\makeatother

\usepackage[pagebackref,breaklinks,colorlinks,citecolor=cvprblue]{hyperref}

\title{{\blendify} -- Python rendering framework for Blender}

\author{
Vladimir Guzov*$^{1,2,3}$
\quad Ilya A. Petrov*$^{1,2}$ \\
\quad Gerard Pons-Moll$^{1,2,3}$ \\
{$^1$\small University of T\"ubingen, Germany\quad $^2$\small T\"ubingen AI Center, Germany} \\
{$^3$\small Max Planck Institute for Informatics, Saarland Informatics Campus, Germany} \\
{\tt\small \{vladimir.guzov, i.petrov\}@uni-tuebingen.de}
}

\begin{document}

\vspace{-30pt}

\twocolumn[{%
    \renewcommand\twocolumn[1][]{#1}%
    \maketitle
    \begin{center}
        \vspace{-2pt}
        \centering
        \begin{overpic}[trim=0cm 0cm 0cm 0cm,clip, width=\linewidth]{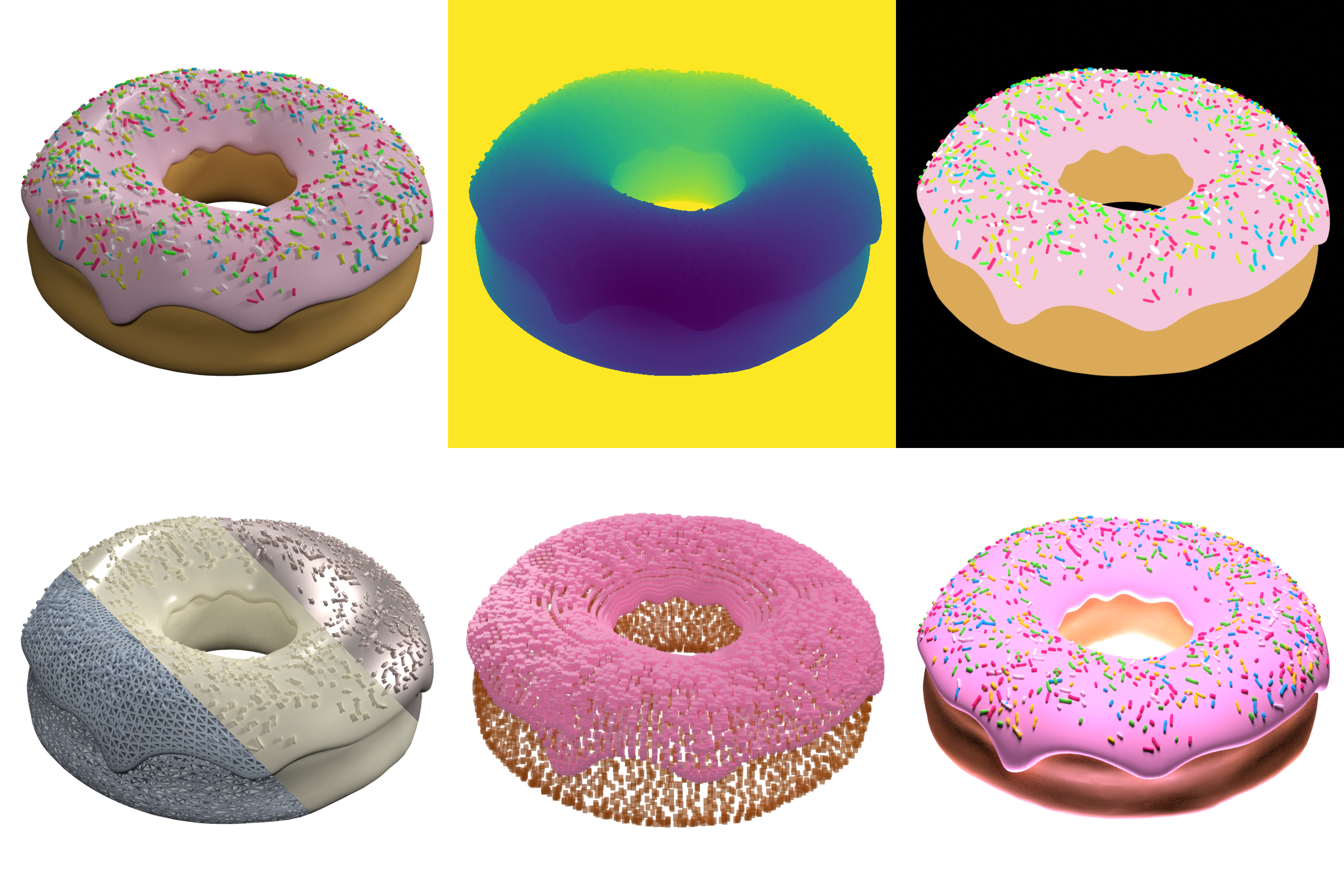}
            \cput(20,650){A) Blender-based rendering}
            \cput(360,650){\color{black}B) Depthmap rendering}
            \cput(690,650){\color{white}C) Albedo rendering}
        
            \cput(20,300){D) Rich material support}
            \cput(360,300){E) Point cloud rendering}
            \cput(690,300){F) Advanced lighting control}
        \end{overpic}
        \vspace{-8pt}
        \captionsetup{type=figure}
        \caption{Various {\blendify} features illustrated on the same donut mesh (for point cloud rendering only the vertices are used). The results of this figure can be reproduced by following the \href{https://virtualhumans.mpi-inf.mpg.de/blendify/walkthrough.html}{walkthrough} on the website and on \href{https://colab.research.google.com/drive/1Y8z52nGkSjxCsJuslprsDflV-lhTz1Hn?usp=sharing}{Google Colab}.}
        \vspace{-2pt}
        \label{fig:features}
    \end{center}%
}]

\begin{abstract}
    \blfootnote{* Equal contribution.}
    With the rapid growth of the volume of research fields like computer vision and computer graphics, researchers require effective and user-friendly rendering tools to visualize results.
    While advanced tools like Blender offer powerful capabilities, they also require a significant effort to master. 
    This technical report introduces {\blendify}, a lightweight Python-based framework that seamlessly integrates with Blender, providing a high-level API for scene creation and rendering. 
    {\blendify} reduces the complexity of working with Blender's native API by automating object creation, handling the colors and material linking, and implementing features such as shadow-catcher objects while maintaining support for high-quality ray-tracing rendering output. 
    With a focus on usability {\blendify} enables efficient and flexible rendering workflow for rendering in common computer vision and computer graphics use cases.
    The code is available at \href{https://github.com/ptrvilya/blendify}{https://github.com/ptrvilya/blendify}.
\end{abstract}

\section{Introduction}
\begin{figure*}[!ht]
    \centering
    \noindent\includegraphics[width=1.0\textwidth]{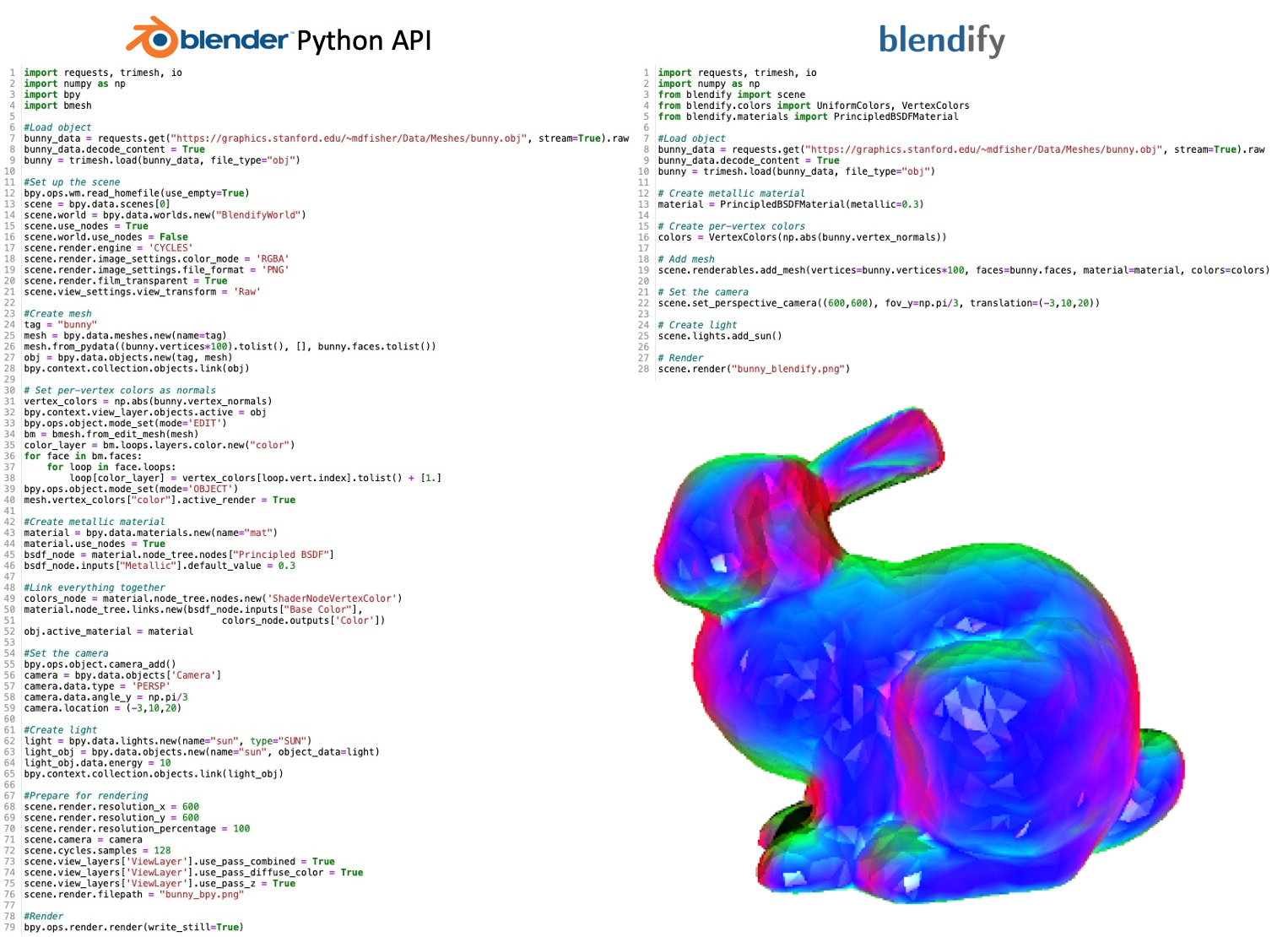}
    \captionsetup{type=figure}
    \caption{Comparison of the code required to render the mesh using native Blender API (on the left) and {\blendify} (on the right).}
    \label{fig:teaser}
    \vspace{2pt}

\end{figure*}

High-quality visualization is not merely an illustration tool but a vital component of analysis and discovery. Along with the constant growth of the number of research articles\footnote{\href{https://arxiv.org/stats/monthly_submissions}{Number of arXiv monthly submissions}}, the need for easy-to-use and effective visualization tools accentuates. 
Visualizations that are accurate while captivating are one of the core components of effective research. They serve not only to demonstrate results but also to communicate vital insights.
Modern research involving 3D models in computer vision and computer graphics, and in other areas, e.g. molecular research, relies on tools that visualize 3D geometry. The last decade saw the development of advanced rendering tools like Blender~\cite{Blender}, Mitsuba~\cite{Mitsuba3} and libraries such as Pytorch 3D~\cite{ravi2020pytorch3d}, Open3D~\cite{zhou2018open3d}, and Pyrender~\cite{matl2018pyrender}. While versatile, these tools also come with a steep learning curve, requiring considerable effort to master. On the other hand, software like Blender is designed to be used via GUI, complicating scripting via Python to automate rendering (e.g., Figure~\ref{fig:teaser} on the left, more than 70 lines of code are needed to generate a simple rendering of a Stanford bunny~\cite{standford1994scans}).

With this technical report, we aim to address these two challenges by introducing a new scientific rendering framework {\blendify} written in Python and based on Blender. {\blendify} is a lightweight Python framework that provides a high-level API for creating and rendering scenes with Blender. Key principles behind the design of {\blendify} are:
\begin{itemize}
    \item ease of use;
    \item seamless integration with Blender's rendering engine;
    \item straightforward automation and assets reuse;
    \item focus on high-quality visualizations for research articles.
\end{itemize}

In the following sections, we outline the main features of the framework (Section~\ref{sec:Features}), describe the underlying architecture (Section~\ref{sec:Architecture}), detail the standalone utilities and algorithms that are part of {\blendify} (Section~\ref{sec:Utilities}), and discuss future directions (Section~\ref{sec:Discussion}).

\section{Features}
\label{sec:Features}

With this section, we overview the key features of {\blendify} and present examples of its application. The example renders in various scenarios are provided in Figure~\ref{fig:features}.
In line with the principles listed above {\blendify} enables:
\begin{itemize}
    \item Export to and import from the Blender \code{*.blend} files;
    \item Rendering depthmap and albedo;
    \item Native support for point cloud rendering, including per-point colored clouds;
    \item Advanced colorization support -- uniform, per-vertex colors, in-memory and file textures, with per-vertex and per-face UV map definitions;
    \item Complex materials that can be defined for any subset of faces on the mesh;
    \item Compatibility with Google Colab~\cite{colab} -- an example notebook is provided \href{https://colab.research.google.com/github/ptrvilya/blendify/blob/main/examples/ipynb/blendify_colab_demo.ipynb}{here}.
\end{itemize}

\noindent The detailed walkthrough of the features with code examples is available on {\blendify} \href{https://virtualhumans.mpi-inf.mpg.de/blendify/walkthrough.html}{website} and on \href{https://colab.research.google.com/drive/1Y8z52nGkSjxCsJuslprsDflV-lhTz1Hn?usp=sharing}{Google Colab}.

\section{Architecture}
\label{sec:Architecture}
\begin{figure*}[!h]
    \vspace{7pt}
    \centering
    \includegraphics[width=1.0\linewidth]{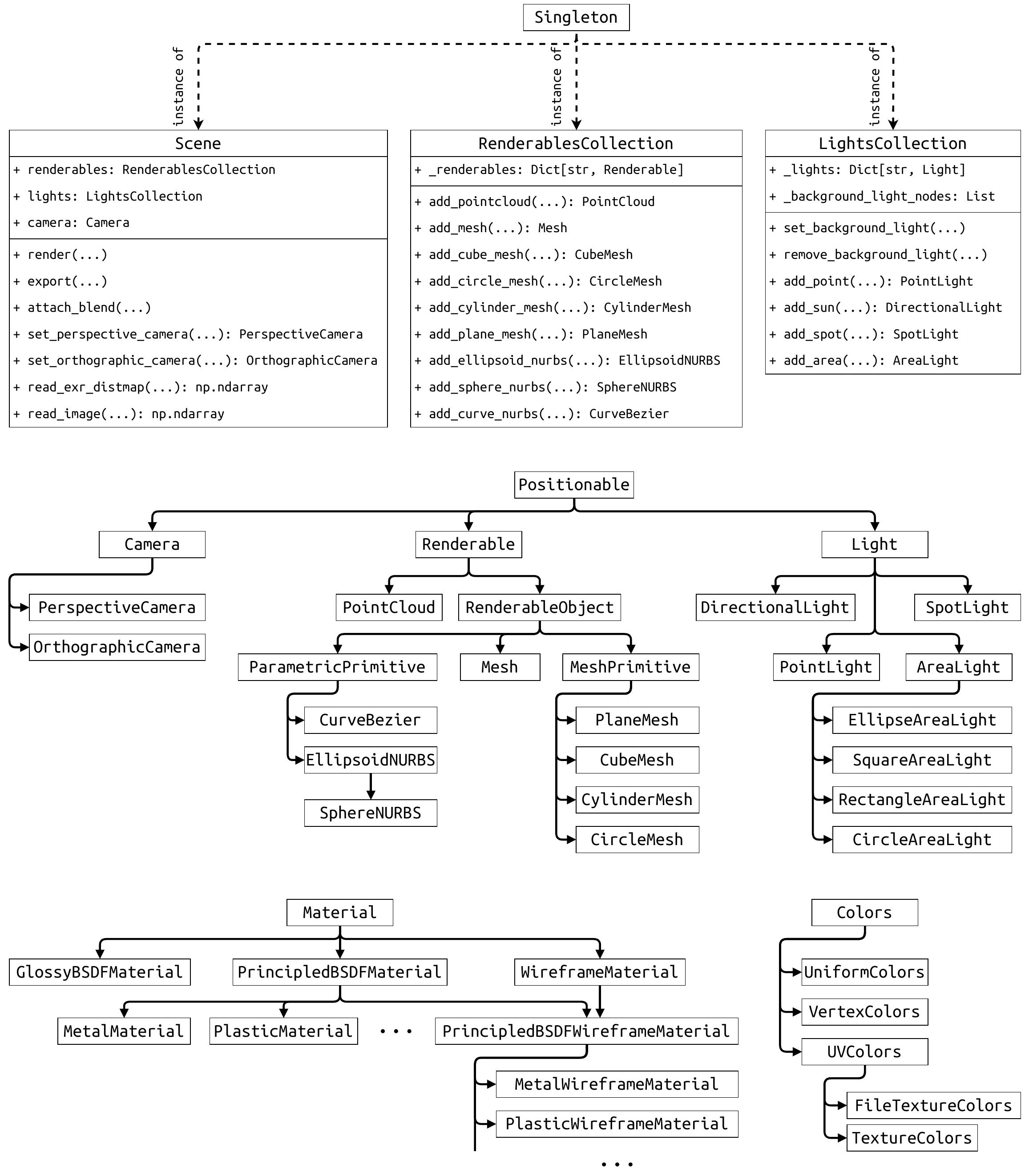}
    \captionsetup{type=figure}
    \caption{Diagram of inheritance for selected classes in {\blendify}.}
    \vspace{7pt}
    \label{fig:architecture}
\end{figure*}

This section documents the development and functionality of {\blendify}. The project aims to simplify the process of scene composition in Blender by providing flexible API for populating the scene and controlling its elements, thereby enabling more complex scene setups with minimal manual intervention. It leverages the Blender Python API (\code{bpy}) to facilitate the creation, modification, and management of various scene elements such as lights, cameras, and materials. 

We provide the inheritance diagram of selected classes in Figure~\ref{fig:architecture}. The core of {\blendify} is the singleton class \code{blendify.scene} that abstracts the corresponding Blender scene and provides functionality to populate it with objects. The \code{scene} class encapsulates collections of 3D objects and light sources that constitute the Blender scene: \code{RenderablesCollection} and \code{LightsCollection}, and stores a single camera for rendering. In the following subsections, we define these collections and their underlying concepts.

\subsection{Scene}
\label{sec:Scene}
The \code{blendify.scene} singleton class stores all the scene objects and implements necessary operations, namely:
\begin{itemize}
    \item camera setup;
    \item rendering;
    \item export to and import from Blender \code{*.blend} files.
\end{itemize}

The export to Blender \code{*.blend} files is implemented via \code{bpy.ops.wm.save\_as\_mainfile} function, thus allowing to save all objects created with {\blendify} (e.g., meshes, materials, primitives, etc.) to be saved as corresponding Blender objects in the file. Due to the complexity of the implementation, import of Blender \code{*.blend} files supports only limited parsing of objects, i.e. only lights and camera from the file can be parsed into {\blendify} objects, while rest of the file's content is appended to the Scene as it is and is not parsed into internal structures.

To unify operations with all the objects that populate the scene, we define an abstract class \code{Positionable}, which all objects inherit from. 
This class implements basic operations to set the object's global position in the scene through interacting with Blender API, tagging it, and handling its destruction. 
The implementation supports all common ways to set the rotation, namely quaternion, axis-angle, rotation matrix, and Euler angles. Additionally, we implement the \code{look\_at} method to define rotation by a specified point to look at.
Geometry objects (meshes, point clouds, primitives), Lights, and Camera are all build on top of \code{Positionable}. The further sub-sections are detailing implementations of these objects.

\subsection{Renderables}
The common abstract parent class for all the 3D geometrical entities that can be rendered is \code{Renderable(Positionable)} defined in \code{blendify.renderables}. This class defines a common interface to update object's visuals (i.e. material and color). As implementing point clouds with Blender structures is tricky further unification can only be done for objects represented with meshes. This is done via another abstract class \code{RenderableObject(Renderable)}, which implements routines related to materials and colors for meshes and primitives.

Scene stores all the \code{Renderable} objects in {scene.renderables} a collection implemented by \code{blendify.renderables.RenderablesCollection}. This singleton class encapsulates a Python dictionary and implements user interface to add \code{Renderable}'s to the scene.  

Further paragraphs detail all renderables implemented in {\blendify}: point clouds, meshes, and primitives.

\PAR{Point clouds}

\textbf{Implemented with:} \code{PointCloud(Renderable)}.

\textbf{Details:} We use implementation provided by Blender Photogrammetry Importer \cite{bullinger2020photogrammetry} as Blender does not yet have a native support for point clouds as geometrical objects. This implementation stores point cloud's vertices as \code{ParticleSystem} object from Blender.

\textbf{Features:} Point clouds support per-point and uniform coloring. Additionally, various types of primitives to represent points are available including cubes, and spheres. One more adjustable feature is particle emission strength that enhances realism by smoothing the light distribution around the point cloud.

\textbf{Limitations:} The implementation through \code{ParticleSystem} restricts the possible options to color the \code{PointCloud} objects. Currently only uniform and per-point coloring is supported, while textured point clouds are not supported (more details on this are given in Section~\ref{sec:ColorsMaterials}).

\PAR{Meshes}

\textbf{Implemented with:}  \code{Mesh(RenderableObject)}.

\textbf{Details:} The object is initialized with vertices and faces that define the desired geometry. 

\textbf{Features:} Meshes can be colored with all currently supported methods: uniformly, per-vertex, and with textures. Moreover {\blendify} supports per-face definition of materials to allow for more realistic renderings (more details are provided in Section~\ref{sec:ColorsMaterials}).

\PAR{Primitives}

\textbf{Implemented with:} mesh-based \code{MeshPrimitive(RenderableObject)} and parametric-based \code{ParametricPrimitive(RenderableObject)}.

\textbf{Details:} Each primitive corresponds to a Blender primitive from \code{bpy.ops}.

Mesh-based primitives include:
\begin{itemize}
    \item \code{CubeMesh(MeshPrimitive)}
    \item \code{CircleMesh(MeshPrimitive)}
    \item \code{CylinderMesh(MeshPrimitive)}
    \item \code{PlaneMesh(MeshPrimitive)}
\end{itemize}

Parametric-based primitives include:
\begin{itemize}
    \item \code{EllipsoidNURBS(ParametricPrimitive)}
    \item \code{SphereNURBS(EllipsoidNURBS)}
    \item \code{CurveBezier(ParametricPrimitive)}
\end{itemize}

\textbf{Features:} The primitives share the same parameters as their Blender counterparts (e.g. size, number of vertices for mesh-based and radius for the parametric-based). Moreover, the \code{PlaneMesh} can serve as a shadow catcher object, i.e. contributing only shadows that are cast on its surface to the final rendered result.

\textbf{Limitations:} Primitive objects support a uniform coloring and a single material per instance. Mesh-based primitives additionally support per-face color and material assignment.

\subsection{Colors and materials}
\label{sec:ColorsMaterials}
For finer control of visuals, \blendify{} uses a combination of materials and colors. 
Types of colors determine the way they are applied to mesh.

\textbf{UniformColors} applies one RGB or RGBA color to the entire mesh or pointcloud.

\textbf{VertexColors} applies a color at each vertex of the mesh or each point of the pointcloud. Despite the same class being used for both pointcloud and meshes, the coloring procedures are very different. Therefore, the class itself contains only the metadata required to apply to color, and each renderable class implements its own coloring algorithm in \code{\_blender\_set\_colors} 
method. 

In the case of the mesh, \code{bmesh} is used to access each face and assign a color to each vertex within a face. During rendering, each face is filled with a barycentric interpolation of colors assigned to vertices. 

In the case of the pointcloud, all the colors are mapped to a texture, one pixel per color, and each point in the cloud is assigned a UV coordinate corresponding to its color.

\textbf{TextureColors} implements texturing for meshes. The texture can only be assigned to the mesh if a position on the mesh is known for each texture pixel. This is usually done using UV coordinate maps, which can be implemented in two ways: either assigning a texture coordinate for each vertex of the mesh or assigning a position for vertices within each face. The former is simpler to use and understand, while the latter offers more flexibility since one mesh vertex can be assigned to several UV coordinates depending on how many faces it is involved in. \blendify{} implements both strategies with \code{VertexUV} and \code{FacesUV} respectively. Similar to \code{VertexColors}, the color of each point on the mesh is then determined by barycentric UV-coordinate interpolation within a face. Additionally, a texture can be accessed from the hard drive without the need to preload it to the memory using \code{FileTextureColors}. This can be handy if a very large texture needs to be used, e.g., a human body movement example in our repository features a scene texture with a resolution of $30,000 \times 30,000$ pixels.

\PAR{Materials}
Classes that implement two basic BSDF based materials are: \code{PrincipledBSDFMaterial}, \code{GlossyBSDFMaterial}. These classes internally create a corresponding Blender shading node that implements the material.
\code{PrincipledBSDFMaterial} is versatile and can be adjusted to approximate a lot of materials. For instance, {\blendify} implements \code{MetalMaterial} and \code{PlasticMaterial} that simply define parameters of the BSDF to approximate corresponding materials.

More complex materials are implemented as a combination of shading nodes.
\code{WireframeMaterial} is an abstract class that implements method \code{overlay\_wireframe} to add shading nodes generating wireframe on top of any given material.
\code{PrincipledBSDFWireframeMaterial} is a \code{PrincipledBSDFMaterial} with a \code{WireframeMaterial} overlayed on top. \code{PlasticWireframeMaterial} and \code{MetalWireframeMaterial} are implemented similarly to non-wireframe materials by setting the parameters of \code{PrincipledBSDFWireframeMaterial} to pre-defined values.

The design of \code{Material} in {\blendify} allows users to easily extend the range of supported materials. To implement the new material, one needs to redefine the \code{create\_material} method, which creates required shading nodes, connects them, and defines the corresponding inputs (\eg, color, alpha, etc.).

\subsection{Lights}
\label{sec:Lights}
In a similar fashion to the \code{RenderablesCollection}, the \code{blendify.lights} module introduces a \code{LightsCollection} as a Singleton class that manages various types of light sources within a scene. These light sources include background light, point lights, directional lights, spotlights, and area lights. Each type of light comes with customizable properties such as strength, color, and shadow emission settings.

The background light is implemented using \code{ShaderNodeBackground} from Blender, providing uniform ambient lighting for the scene. All other light types are modeled after their corresponding Blender lights and support adjustments for color, strength, size, and other properties.

\subsection{Camera}
\label{sec:Camera}
The framework supports two types of cameras: \code{PerspectiveCamera} and \code{OrthographicCamera}, that mirror the features of the corresponding Blender cameras. Setting up the camera is implemented with two methods of the Scene class: \code{Scene.set\_perspective\_camera} and \code{Scene.set\_orthographic\_camera}. Additionally, the camera can be loaded from the \code{*.blend} file via \code{Scene.attach\_blend\_with\_camera}.

To ease the manual camera setup {\blendify} implements \code{look\_at} rotation mode that directs the camera to the specified point. For that, the forward vector (Z-axis) of the camera is set as a vector pointing from the camera position to the target point. Then an upright direction is determined as a vector aligned with the Z-axis of the world. In case a camera is required to look in the same direction, an upright vector aligns with the Y-axis. Next, the right camera direction (X-axis) is determined by a cross-product of the forward and upright vectors, which is followed by recalculation of an up direction (Y-axis) with a cross-product of forward and right vectors.  

\section{Utilities and algorithms}
\label{sec:Utilities}

\blendify{} also features an additional \code{utils} package, implementing selected tasks met in scientific visualization: 

{\bf\rom{1}. Camera-colored PC.} During the rendering process of sparse pointclouds, the widely used technique to reduce the visual noise is to hide the back-facing points by decolorizing them or increasing their opacity. This requires the user to change the colorization of the points based on the direction of the camera. We ease the creation of such conditionally-colored pointclouds by implementing routines used for approximating the normals based on the neighboring points (\code{estimate\_normals\_from\_pointcloud}) and determining a facing direction and color of each point in a pointcloud based on their normals and camera direction (\code{approximate\_colors\_from\_camera}). The \textit{"Camera colored point cloud"} example in our repository implements this use case by rendering a sparse pointcloud of the Stanford bunny.

{\bf\rom{2}. Camera trajectory interpolation.} One common task during video creation is to gradually move a camera between the specified key positions and rotations, creating a smooth camera movement animation. For that, a \code{Trajectory} class was implemented, containing \code{add\_keypoint} method to set the key  orientation points in time and \code{refine\_trajectory} method to produce a per-frame list of camera positions and rotations, forming a smooth trajectory. Such trajectory refinement can be seen in \textit{"SMPL movement"} example in our repository.

{\bf\rom{3}. Pointcloud to mesh texture transfer.} To render large pointclouds, such as scans, several techniques can be used, one of which is to turn the pointcloud into the textured mesh before rendering. The mentioned method is especially effective with raytracing rendering engines like the one used in \blendify{} because it results in correct light reflections from a mesh surface and faster rendering times due to simplified geometry. While the mesh itself can be formed relatively easily with existing tools (such as Meshlab~\cite{meshlab} or Open3D~\cite{zhou2018open3d}), no easy-to-use tool for texture transfer exist. We are filling this gap by implementing \code{meshify\_pc} function. Given a colored pointcloud, \code{meshify\_pc} performs a mesh creation, geometry simplification and color transfer is 4 steps:
1) Using the ball pivoting algorithm~\cite{bernardini1999ballpivot}, a mesh is formed from the pointcloud. If the pointcloud does not have normals, they are estimated from the nearest neighbor's landscape.
2) The geometry of the mesh is simplified~\cite{garland1997surface} to reduce computations during rendering.
3) A UV-map for a mesh is created with a naïve algorithm: each face is mapped to a separate triangle on the texture, with a small gap between triangles to prevent color spilling. This algorithm does not require mesh unwrapping, which might be a complicated task for large scenes. On the negative side, the necessity to create borders between the triangles reduces the useful space on the texture. 
4) As a last step, each point of the input pointcloud is projected to a mesh, and for each projected point a corresponding UV coordinate on the texture is determined. This information is used to determine color of each texture pixel with weighed average of the nearest projected points' colors.
The resulting textured mesh can then be used in \blendify{} for rendering.

\section{Discussion \& Future work}
\label{sec:Discussion}
{\blendify} is a flexible visualization framework designed with a focus on visualization for scientific articles in the field of computer vision and computer graphics. The implementation deliberately sacrifices a lot of Blender features, such as native animation support and physics modeling, to ease the interaction with the framework.   

One of the promising directions of developing {\blendify} is to integrate Point Cloud developed by Blender (coming in future releases)\footnote{\href{https://docs.blender.org/manual/en/4.1/modeling/point_cloud.html}{https://docs.blender.org/manual/en/4.1/modeling/point\_cloud.html}} instead of currently used Blender Photogrammetry Importer \cite{bullinger2020photogrammetry}. This change will potentially allow us to simplify the materials and colors implementation for point clouds and unify it with the \code{Mesh} class. 

The material support can also be improved by adding an ability to control material properties (metallic, roughness, glossiness, \etc) with a texture -- such property encoding is sometimes used in materials for complex 3D models.

Another possible way to improve {\blendify} is to enable more comprehensive parsing of \code{*.blend} files into internal structures to allow interactive modification of objects with import and export to a file in between.

\section{Conclusions}
\label{sec:Conclusions}
In this technical report we presented {\blendify} -- a Python rendering framework for scientific visualization based on Blender. The framework focuses on the common use cases in computer vision and computer graphics. {\blendify} provides an intuitive, high-level interface that simplifies scene composition, rendering, and material management. The framework features support for all common geometry types, coloring options, and materials. 
{\blendify} democratizes access to sophisticated rendering tools, empowering researchers to more easily integrate high-quality visualizations into their work. Future developments, such as improved point cloud support and expanded import capabilities, will continue to enhance Blendify’s functionality, ensuring it remains a useful tool for the scientific community.

\small
\paragraph{Acknowledgements}
Special thanks to István Sarandi and Riccardo Marin for their help in testing the code and suggesting the features for the framework. We also thank RVH group members for their feedback that helped to improve Blendify.
The project was made possible by funding from the Carl Zeiss Foundation.
This work is supported by the Deutsche Forschungsgemeinschaft (DFG, German Research Foundation) - 409792180 (EmmyNoether Programme, project: Real Virtual Humans) and the German Federal Ministry of Education and Research (BMBF): Tübingen AI Center, FKZ: 01IS18039A. 
G. Pons-Moll is a member of the Machine Learning Cluster of Excellence, EXC number 2064/1 – Project number 390727645. 
The authors thank the International Max Planck Research School for Intelligent Systems (IMPRS-IS) for supporting I.A. Petrov.

{
    \small
    \bibliographystyle{ieeenat_fullname}
    \bibliography{main}
}

\end{document}